\documentclass[11pt,a4paper]{article}
\usepackage{times,latexsym}
\usepackage{url}
\usepackage[T1]{fontenc}
\usepackage[T1]{CJKutf8}

\usepackage[acceptedWithA]{tacl2021v1}

\usepackage{xspace,mfirstuc,tabulary}

\newif\iftaclinstructions
\taclinstructionsfalse 
\iftaclinstructions
\renewcommand{\confidential}{}
\renewcommand{\anonsubtext}{(No author info supplied here, for consistency with
TACL-submission anonymization requirements)}
\newcommand{\instr}
\fi

\iftaclpubformat 

\else

\fi

\usepackage{graphicx}

\usepackage{color}

\usepackage{bm}

\usepackage{amssymb,amsmath}
\usepackage{amsfonts} 
\usepackage{bbm}
\usepackage{colortbl}
\usepackage{subcaption}

\title{Average Token Delay: A Duration-aware Latency Metric for Simultaneous Translation}

\author{
  Yasumasa Kano$^\diamond$ 
  \and
  Katsuhito Sudoh$^\diamond$
  \and
  Satoshi Nakamura$^\diamond$
  \\
  \ \\
  $^\diamond$Nara Institute of Science and Technology / 8916-5 Takayama-cho, Ikoma, Nara, Japan
  \\
  \texttt{kano.yasumasa.kw4@is.naist.jp}
}

\date{}

\begin{document}
\begin{CJK}{UTF8}{min}

\maketitle
\begin{abstract}
Simultaneous translation is a task in which the translation begins before the end of an input speech segment.
Its evaluation should be conducted based on latency in addition to quality,
and for users, the smallest possible amount of latency is preferable.
Most existing metrics measure latency based on the start timings of partial translations and ignore their duration.
This means such metrics do not penalize the latency caused by long translation output, which delays the comprehension of users and subsequent translations.
In this work, we propose a novel latency evaluation metric for simultaneous translation called \emph{Average Token Delay} (ATD) that focuses on the duration of partial translations.
We demonstrate its effectiveness through analyses simulating user-side latency based on Ear-Voice Span (EVS).
In our experiment, ATD had the highest correlation with EVS among baseline latency metrics under most conditions.
\end{abstract}

\section{Introduction}\label{sec:introduction}
Machine translation (MT), which has evolved rapidly due to recent neural network techniques, is now widely used for both written and spoken languages.
MT for speech attracts the translation of various conversations, lecture, talks, etc.
In situations requiring real-time communication, MT should function simultaneously 
without waiting for the speech's conclusion.
Such a task is often called \emph{simultaneous machine translation} (SimulMT).
Hereinafter,  the term SimulMT covers both text and speech inputs.
One crucial challenge in SimulMT is the quality-latency trade-off.
Although taking a longer input with a longer wait can improve the translation quality, unfortunately it also creates more latency, and \textit{vice versa}.

\begin{figure}[t]
\centering
\centerline{\includegraphics[width=8.0cm]{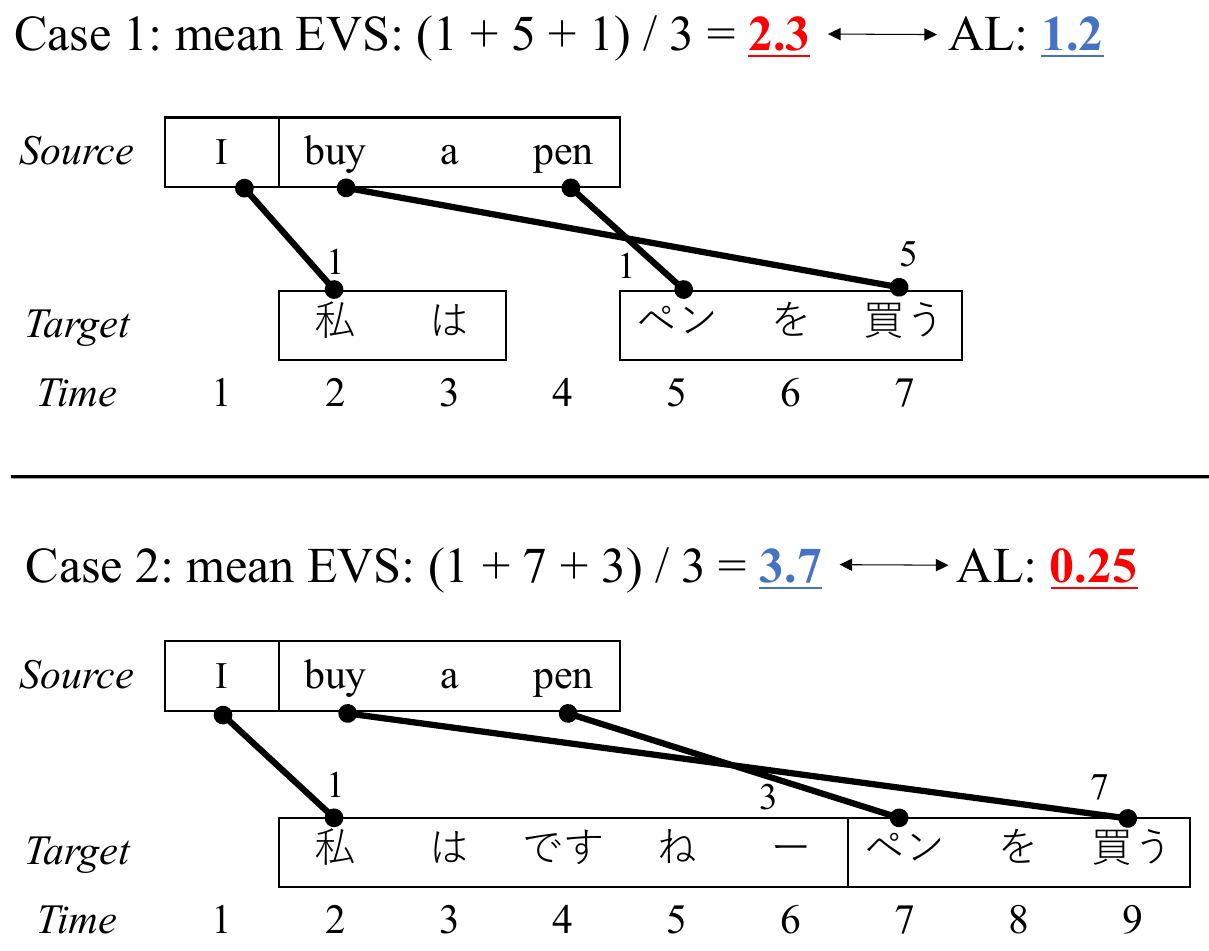}}
\caption{Example calculation of mean Ear-Voice Span (EVS): Black lines represent word alignment, and time distance of the lines is EVS.
This example excludes word alignment for stop words and function words.
The mean EVS is smaller in case 1; AL is smaller in case 2.}
\label{fig:EVS_AL}
\end{figure}

In SimulMT studies, we compare models in such a quality-latency trade-off.
In the research of human interpretation, Ear-Voice Span (EVS) has been used to evaluate latency.
Basically, EVS is the time difference between the source word and the corresponding target word, although variations can be found \cite{elisa2019}.

\autoref{fig:EVS_AL} illustrates an example calculation of a mean EVS.
For both cases, the first input chunk is translated into the first output chunk, and the second one is translated next.
In this example, EVS is represented as the time distance of the black lines connecting the source and the semantically corresponding target word.
When we calculate the mean value of these EVSs, case 1 has a smaller value.
Therefore, the latency of case 1 is smaller than that of case 2.

Although EVS is a persuasive latency metric, it has not been used in SimulMT research.
One reason is that SimulMT research emerged from text-to-text simultaneous translation while EVS is a latency metric for speech-to speech simultaneous translation by human interpreters.
Therefore, latency metrics have been proposed that can calculate the latency of text-to-text simultaneous machine translation.

One example is Average Lagging \citep[AL;][]{ma-etal-2019-stacl}, which is the most commonly used latency metric.
AL, which is based on the number of input words that are available when starting a translation, measures the average amount of input words over the whole translation process.
It matches wait-$k$ SimulMT \citep{ma-etal-2019-stacl} that waits for the $k$ input tokens before starting the translation and alternately repeats reading and writing one token.
In \autoref{fig:EVS_AL}, AL is smaller in case 2, which does not agree with its mean EVS.
This difference comes from the definition of AL, which counterintuitively gives a smaller latency value to a long chunk output at a time step.
Such a long chunk output causes inevitable delays for the translation of subsequent parts in speech-to-speech SimulST because the speech synthesis cannot start a new speech output while \emph{speaking}.
Even for closed caption output, long outputs in a short time period cause a high cognitive load for users.
Therefore, since a translation output's length affects user experiences, this length should be considered in latency measurements.
This observation suggests the need for another latency metric to cope with such situations.

In this work, we propose a novel latency metric called \emph{Average Token Delay} (ATD)\footnote{ATD is implemented in \url{https://github.com/facebookresearch/SimulEval}.} that focuses on the end timings of partial translations in SimulMT.
ATD generalizes latency measurements for both speech and text outputs and works intuitively for chunk-based outputs that are not properly handled by AL, as discussed above.
ATD is much simpler than EVS, which requires a long process like transcribing input and output speech, getting word timestamps, and obtaining word alignment by annotators.
We present simulation results that clarify ATD's characteristics and
demonstrate its effectiveness through SimulMT experiments and comparisons with ATD with baseline latency metrics.
Since no experiments have compared latency metrics, no evaluation dataset exists.
Therefore, we created a small dataset and used it to compare the correlation between each latency metric and the mean EVS.
In our experiments, ATD had the highest correlation among the baseline latency metrics in most conditions.

\section{Simultaneous Machine Translation}
First, we review the SimulMT formulation to share the notation used in this paper.

In standard sentence-level NMT,
let $\bm{x} = x_1,x_2,...,x_m$ be an input sentence and $\bm{y} = y_1,y_2,...,y_n$ be its translation,
the output probability is denoted as
\begin{equation}
p(\bm{y}|\bm{x}) = \prod_{t=1}^{n} P(y_t | \bm{x}, \bm{y}_{<t}).
\end{equation}

SimulMT takes a prefix of the input for its incremental decoding:
\begin{equation}
p(\bm{y}|\bm{x}) = \prod_{t=1}^{n} P(y_t | \bm{x}_{\leq g(t)},\bm{y}_{<t}),
\end{equation}
where $g(t)$ is a monotonic non-decreasing function that represents the number of input tokens read until the prediction of $t$-th output token $y_t$, so that $\bm{x}_{\leq g(t)}$ means an input prefix, and $\bm{y}_{<t}$ is the prefix translation predicted so far.
This means that we can obtain a pair of an input prefix and a corresponding output prefix $(\bm{x}_{\leq g(t)}, \bm{y}_{\leq t})$ by that time.

The incremental decoding can be represented by a sequence of actions, \textit{READ} and \textit{WRITE}.
\textit{READ} is an action that takes a new input, typically one token for the text input or a fixed number of frames for the speech input.
\textit{WRITE} is an action that predicts an output, typically one token for a text output or a corresponding speech signal for the speech output.

\section{Existing Latency Metrics for Simultaneous Translation}
Several latency metrics have been proposed in the SimulMT field.
In this section, we review them before we propose ATD.

\citet{gu-etal-2017-learning} proposed a latency metric called the Consecutive Wait length (CW), which
counts the number of consecutive waited input tokens between two output tokens and measures the local delays in a sentence.
\citet{cho-2016} proposed an AP that measures the average latency for an entire sentence.
However, AP suffers from the following problem: the latency value differs depending on the input and the output sequence lengths even for the same READ-WRITE strategy.

\citet{ma-etal-2019-stacl} proposed Average Lagging (AL), which has recently become the most commonly used method,
denoted as follows:
\begin{equation}
AL_g(\bm{x}, \bm{y}) = \frac{1}{\tau_g(|\bm{x}|)} \sum_{t =1}^{\tau_g(|\bm{x}|)} \left( g(t) - \frac{t -1}{r} \right),
\label{eqn:AL_g}
\end{equation}
where $r$ is the length ratio defined as $|\bm{y}|/|\bm{x}|$ and $\tau_g(|\bm{x}|)$ is the cut-off step:
\begin{equation}
\tau_g(|\bm{x}|) = \min\{t \mid g(t) = |\bm{x}|\},
\label{eqn:tau_g}
\end{equation}
denoting the index of the output token predicted right after the observation of the entire source sentence.

However, AL still suffers from unintuitive latency measurement because it can be negative when the model finishes the translation before reading the entire input.
This is because if $|\bm{y}| << |\bm{x}|$, then $r << 1$, and so the second term $(t-1)/r$ in the subtraction becomes too large.
To mitigate this problem, \citet{ma-etal-2020-simuleval} modified AL by changing the calculation of length ratio $r$ to $|\bm{y^*}|/|\bm{x}|$ based on the length of reference translation $\bm{y^*}$.
\citet{papi-etal-2022-generation} proposed Length-Adaptive Average Lagging (LAAL), which modified $r$ to $max(|\bm{y}|,|\bm{y^*}|)/|\bm{x}|$ to appropriately evaluate the latency of a translation that is longer than the reference.
However, these modifications are not enough to deal with the problem of AL in which the latency becomes smaller for longer partial output.
When the partial translation output becomes longer, $t$ increases and $(t-1)/r$ becomes larger although $g(t)$ does not change.
As a consequence, the result of subtraction in \autoref{eqn:AL_g} is reduced.

\citet{arivazhagan-etal-2019-monotonic} proposed another AL variant called Differentiable Average Lagging (DAL), which can optimize a simultaneous translation model:

\begin{equation}
DAL_g(\bm{x}, \bm{y}) = \frac{1}{|\bm{y}|} \sum_{t =1}^{|\bm{y}|} \left( g'(t) - \frac{t -1}{r} \right),
\label{eqn:DAL_g}
\end{equation}

\begin{equation}
\label{eqn:DAL_g_dash}
g'(t) = \begin{cases} g(t) &  t = 1 \\
\max(g(t), g'(t - 1) + r) & t > 1 \end{cases}.
\end{equation}

DAL replaces the $g(t)$ of AL with $g'(t)$ and does not use cut-off step $\tau_g(|\bm{x}|)$.
DAL mitigates the AL problem that long chunk output reduces the delay, although it remains insufficient.
Suppose $|\bm{x}| = |\bm{y}|$ ($r = 1$).
When the partial translation output becomes longer than the partial input, $g'(t-1)+r$ exceeds $g(t)$ in Equation~\ref{eqn:DAL_g_dash}.
In this situation, every time a new target token is output, $g'(t)$ increases by one, as does $(t-1)/r$.
Therefore, the difference between the two terms in Equation~\ref{eqn:DAL_g} is not changed by long output; nor does the delay increase.
Even though a long output should delay the start of the next chunk translation, such delay is not counted in DAL.

The above latency metrics are proposed to evaluate sentence-level SimulMT, and so \citet{iranzo-sanchez-etal-2021-stream-level} proposed a method to calculate them for streaming input by converting the global delay in the streaming input to the local delay of each sentence.

\section{Proposed Metric: Average Token Delay}
We propose a novel latency metric ATD to include the delay caused by a long chunk output in latency measurement.
We start from the ATD calculation in the case of speech-to-speech SimulMT and generalize it for speech-to-text and text-to-text cases.

In the following explanation, we first suppose the computation time is included in ATD's latency calculation.
Then we explain the non-computation-aware version of ATD which excludes the computation time.

\begin{figure}[t]
\centering
\centerline{\includegraphics[width=8.0cm]{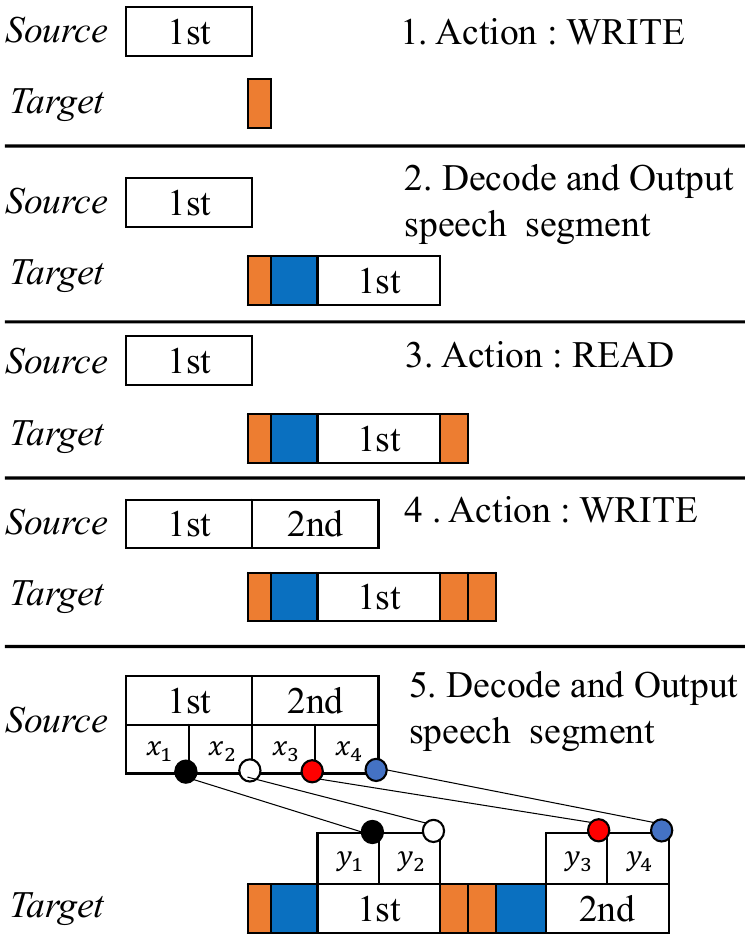}}
\caption{Step-by-step example of simultaneous speech-to-speech MT: Time passes from left to right.}
\label{fig:s2s_simulate}
\end{figure}

\subsection{ATD for simultaneous speech-to-speech translation}
\autoref{fig:s2s_simulate} illustrates a step-by-step workflow of a speech-to-speech SimulMT.
In this figure, the white boxes with an order (1st, 2nd) represent the duration of the speech segments, the orange ones represent the processing time to encode the input prefixes and to judge whether we should \emph{READ} or \emph{WRITE}, and the blue ones represent the decoding time.

To calculate ATD, we divide each speech segment into sub-segments of length $\tau$ from the beginning of the segment, assuming one word is uttered in duration $\tau$.
Intuitively, ATD is the average of the time differences between the points of the same color (black, white, red, and blue) at the ending time of sub-segments (Step 5).

ATD is defined as follows:
\begin{equation}
\textrm{ATD}(\bm{x}, \bm{y})= \frac{1}{|\bm{y}|}\sum_{t=1}^{|\bm{y}|} \left( T(y_{t}) - T(x_{a(t)}) \right)
\label{eqn:atd_definition},
\end{equation}
where
\begin{equation}
\label{eqn:a_definition}
a(t) = \min(t - d(t), g(t))
\end{equation}
\begin{equation}
\label{eqn:d_definition}
d(t) = (t-1) - a(t-1).
\end{equation}

$T(\cdot)$ in \autoref{eqn:atd_definition} represents the ending time of each input or output token, shown as colored points in \autoref{fig:s2s_simulate}.
The token is a sub-segment in speech; it is a character or a word in text.
$a(t)$ represents the index of the input token corresponding to $y_{t}$ in the time difference calculation and $a(0)$ = 0.
$d(t)$ in \autoref{eqn:d_definition} represents how much longer the duration of the previous translation prefix is than that of the previous input prefix.
As shown in \autoref{eqn:a_definition}, if $d(t)$ > $0$, $a(t)$ becomes smaller than output index $t$.
This means the previous long output increases the time difference between the input and corresponding output tokens.

ATD is the average delay of the output tokens against their corresponding input tokens, considering the latency required for inputs and outputs.
Although the input-output correspondence does not necessarily denote semantic equivalence, especially for language pairs with large differences in their word order and numbers of tokens, we use this simplified formulation for latency measurements like in AL.

\begin{figure}[t]
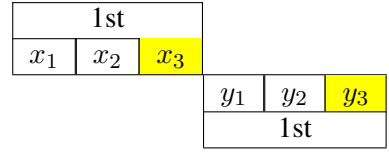
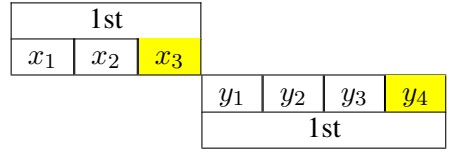

\centering

\begin{minipage}{1.0\hsize}
\centering
\begin{tabular}{cccccc}
\cline{1-3}
\multicolumn{3}{|c|}{1st} & & & \\
\cline{1-3}
\multicolumn{1}{|c|}{$x_1$} & \multicolumn{1}{|c|}{$x_2$} & \multicolumn{1}{|c|}{\cellcolor[rgb]{1.0, 1.0, 0.0}{$x_3$}} & & & \\ \hline
& & & \multicolumn{1}{|c|}{$y_1$} & \multicolumn{1}{|c|}{$y_2$} & \multicolumn{1}{|c|}{\cellcolor[rgb]{1.0, 1.0, 0.0}$y_3$} \\ \cline{4-6}
& & & \multicolumn{3}{|c|}{1st} \\ \cline{4-6}
\end{tabular}
\subcaption{ 1st translation has identical length as 1st input}
\label{tab:equation1a}
\end{minipage}

\vspace{3mm}

\begin{minipage}{1.0\hsize}
\centering
\begin{tabular}{ccccccc}
\cline{1-3}
\multicolumn{3}{|c|}{1st} & & & & \\
\cline{1-3}
\multicolumn{1}{|c|}{$x_1$} & \multicolumn{1}{|c|}{$x_2$} & \multicolumn{1}{|c|}{\cellcolor[rgb]{1.0, 1.0, 0.0}$x_3$} & & & & \\ \hline
& & & \multicolumn{1}{|c|}{$y_1$} & \multicolumn{1}{|c|}{$y_2$} & \multicolumn{1}{|c|}{$y_3$} & \multicolumn{1}{|c|}{\cellcolor[rgb]{1.0, 1.0, 0.0}$y_4$} \\ \cline{4-7}
& & & \multicolumn{4}{|c|}{1st} \\ \cline{4-7}
\end{tabular}
\subcaption{1st translation is longer than 1st input}
\label{tab:equation1b}
\end{minipage}
\caption{Examples of 1st chunk translation}
\label{tab:equation1}
\end{figure}

\begin{figure}[t]
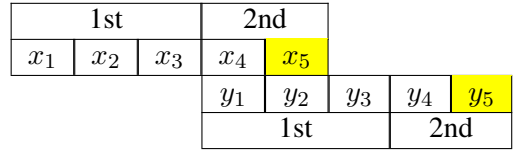
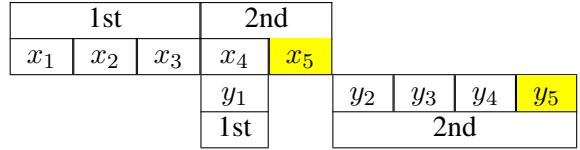
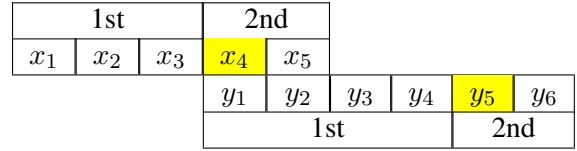

\centering

\begin{minipage}{1.0\hsize}
\centering
\begin{tabular}{cccccccc}
\cline{1-5}
\multicolumn{3}{|c|}{1st} & \multicolumn{2}{|c|}{2nd} & & & \\
\cline{1-5}
\multicolumn{1}{|c|}{$x_1$} & \multicolumn{1}{|c|}{$x_2$} & \multicolumn{1}{|c|}{$x_3$} & \multicolumn{1}{|c|}{$x_4$} & \multicolumn{1}{|c|}{\cellcolor[rgb]{1.0, 1.0, 0.0}$x_5$} & & & \\ \hline
& & & \multicolumn{1}{|c|}{$y_1$} & \multicolumn{1}{|c|}{$y_2$} & \multicolumn{1}{|c|}{$y_3$} & \multicolumn{1}{|c|}{$y_4$} & \multicolumn{1}{|c|}{\cellcolor[rgb]{1.0, 1.0, 0.0}$y_5$} \\ \cline{4-8}
& & & \multicolumn{3}{|c|}{1st} & \multicolumn{2}{|c|}{2nd} \\ \cline{4-8}
\end{tabular}
\subcaption{1st translation is same length as 1st input.}
\label{tab:equation2a}
\end{minipage}

\vspace{3mm}

\begin{minipage}{1.0\hsize}
\centering
\begin{tabular}{ccccccccc}
\cline{1-5}
\multicolumn{3}{|c|}{1st} & \multicolumn{2}{|c|}{2nd} & & & & \\
\cline{1-5}
\multicolumn{1}{|c|}{$x_1$} & \multicolumn{1}{|c|}{$x_2$} & \multicolumn{1}{|c|}{$x_3$} & \multicolumn{1}{|c|}{$x_4$} & \multicolumn{1}{|c|}{\cellcolor[rgb]{1.0, 1.0, 0.0}$x_5$} & & & & \\ \hline
& & & \multicolumn{1}{|c|}{$y_1$} & & \multicolumn{1}{|c|}{$y_2$} & \multicolumn{1}{|c|}{$y_3$} & \multicolumn{1}{|c|}{$y_4$} & \multicolumn{1}{|c|}{\cellcolor[rgb]{1.0, 1.0, 0.0}$y_5$} \\ \cline{4-4} \cline{6-9}
& & & \multicolumn{1}{|c|}{1st} & & \multicolumn{4}{|c|}{2nd} \\ \cline{4-4} \cline{6-9}
\end{tabular}
\subcaption{1st translation is shorter than 1st input.}
\label{tab:equation2b}
\end{minipage}

\vspace{3mm}

\begin{minipage}{1.0\hsize}
\centering
\begin{tabular}{ccccccccc}
\cline{1-5}
\multicolumn{3}{|c|}{1st} & \multicolumn{2}{|c|}{2nd} & & & & \\
\cline{1-5}
\multicolumn{1}{|c|}{$x_1$} & \multicolumn{1}{|c|}{$x_2$} & \multicolumn{1}{|c|}{$x_3$} & \multicolumn{1}{|c|}{\cellcolor[rgb]{1.0, 1.0, 0.0}$x_4$} & \multicolumn{1}{|c|}{$x_5$} & & & & \\ \hline
& & & \multicolumn{1}{|c|}{$y_1$} & \multicolumn{1}{|c|}{$y_2$} & \multicolumn{1}{|c|}{$y_3$} & \multicolumn{1}{|c|}{$y_4$} & \multicolumn{1}{|c|}{\cellcolor[rgb]{1.0, 1.0, 0.0}$y_5$} & \multicolumn{1}{|c|}{$y_6$} \\ \cline{4-9}
& & & \multicolumn{4}{|c|}{1st} & \multicolumn{2}{|c|}{2nd} \\ \cline{4-9}
\end{tabular}
\subcaption{1st translation is longer than 1st input.}
\label{tab:equation2c}
\end{minipage}

\caption{Examples of 2nd chunk translation}
\label{tab:equation2}
\end{figure}

\autoref{tab:equation1} shows examples to explain \autoref{eqn:a_definition} for the first chunk translation.
Here we simplify the duration of the input and output tokens to the same length.
In \autoref{tab:equation1a}, we measured the token delay on $y_3$.
Here $d(3)= 0$, and then we obtained $a(3) = t-0 = 3$, and so $y_3$ corresponds to $x_3$.
In \autoref{tab:equation1b}, we measured the token delay on $y_4$,
 obtained $d(4) = 0$, and then obtained $a(4) = g(4) = 3$, and so $y_4$ corresponds to $x_3$.
In \autoref{tab:equation1a}, the input and output lengths are identical, as are the corresponding indexes of the input and output tokens.
However, in \autoref{tab:equation1b}, since the output is longer than the input,
we corresponded the $y_3$ and latter tokens like $y_4$ to identical input token $x_3$.

\autoref{tab:equation2} shows examples to explain \autoref{eqn:a_definition} for the second chunk translation.
Suppose we measure the token delay on $y_5$.
In \autoref{tab:equation2a} and \autoref{tab:equation2b}, $d(5) = 0$, and so $a(5) = t-0 = 5$.
Therefore, $y_5$ corresponds to $x_5$.
In \autoref{tab:equation2c}, $d(5) = 4 - a(4) = 1$, according to the the result of \autoref{tab:equation1b}.
Therefore, $a(5) = t - 1= 4$ and $y_5$ corresponds to $x_4$.
Unlike \autoref{tab:equation2a} and \autoref{tab:equation2b}, the first chunk output, which is longer than the first chunk input in \autoref{tab:equation2c}, produces the difference of the corresponding indexes of the input and output tokens (\autoref{tab:equation1b}).
The difference of the indexes is accumulated and passed to the latter tokens (\autoref{tab:equation2c}).

\begin{figure}[t]
\centering
\centerline{\includegraphics[width=4.0cm]{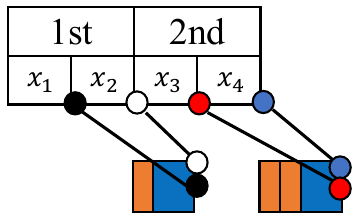}}
\caption{Summary view for latency measurement for simultaneous speech-to-text translation}
\label{fig:s2t_simulate}

\centering
\centerline{\includegraphics[width=2.5cm]{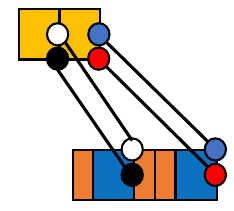}}
\caption{Summary view for latency measurement for simultaneous text-to-text translation}
\label{fig:t2t_simulate}

\centering
\centerline{\includegraphics[width=4.0cm]{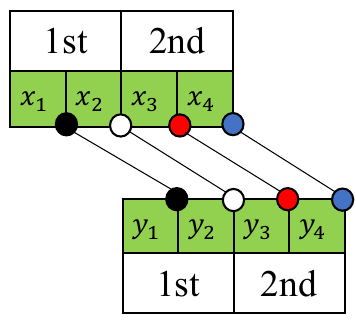}}
\caption{Summary view for non-computation-aware latency measurement for simultaneous text-to-text translation}
\label{fig:t2tNCA_simulate}
\end{figure}

\subsection{ATD for simultaneous \{speech,text\}-to-text translation}
\autoref{fig:s2t_simulate} illustrates the latency measurement for speech-to-text SimulMT, assuming the text output duration is negligible.
\autoref{fig:t2t_simulate} is the latency measurement for the text-to-text SimulMT.
We reserve the input duration here (shown by yellow rectangles) as the computation time of the automatic speech recognition (ASR) because the input for text-to-text SimulMT should come from speech.

\subsection{Non-computation-aware ATD}
We sometimes use a latency measurement independent from the computation time for theoretical analyses.
In Figures~\ref{fig:s2s_simulate}, \ref{fig:s2t_simulate}, and \ref{fig:t2t_simulate}, we remove the orange, blue, and yellow parts and only include the duration of the speech segments to calculate the delay.
However, this means all the terms in the text-to-text SimulMT become 0.
We follow the conventional step-wise latency measurement as AP and AL by letting each input and output word spend one step (\autoref{fig:t2tNCA_simulate}).
Here we assume the model can read the next input and output a partial translation in parallel.

\section{Simulation}
Before experiments using real data,
we show simulations that compare AL, DAL and ATD in different conditions in a text-to-text SimulMT.

\subsection{Cases 1 and 2}\label{subsec:casex}

\begin{table}[t]
\centering
\begin{tabular}{lcccc}
\hline
 & AL & DAL & ATD & EVS \\
\hline
Case 1 & 1.2 & 1.84 & 2.4 & 2.3\\
Case 2 & 0.25 & 1.19 & 3.75 & 3.7\\
\hline
Increase rate & -79\% & -0.35\% & +36\% & +38\% \\
\hline
\end{tabular}
\caption{Latency values for cases in \autoref{fig:EVS_AL}}
\label{tb:case1and2}
\end{table}

First, we calculated the latency metrics for the two cases in \autoref{fig:EVS_AL} and show the calculated values in \autoref{tb:case1and2}.
AL and DAL are larger in case 1, while ATD is larger in case 2 as is EVS.
This example shows that ATD sufficiently considers the delay caused by the long outputs.

\subsection{Case 3: 20-20}\label{subsec:case1}

\begin{figure}[t]
\centering
\centerline{\includegraphics[width=8.5cm]{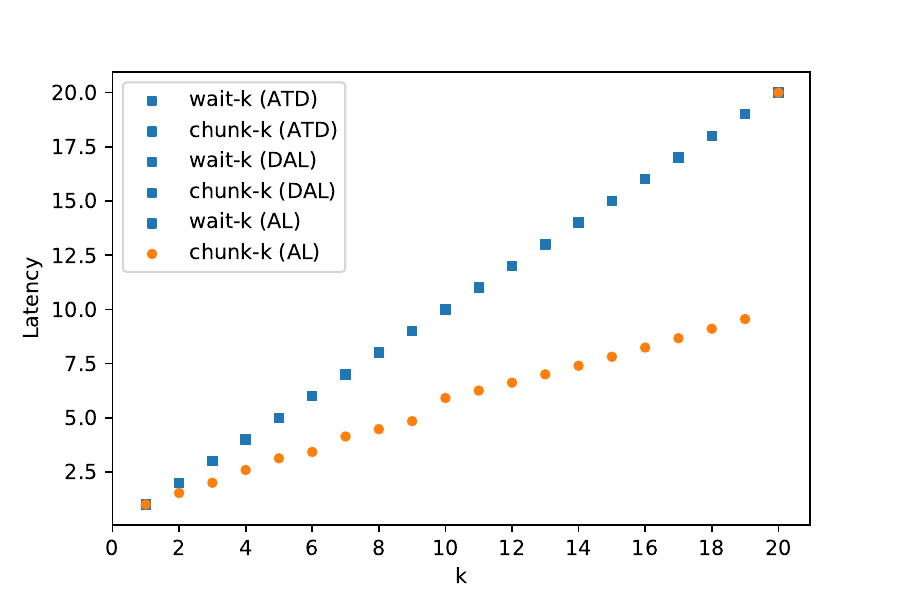}}
\caption{Case 3 (20 input and 20 output tokens)}
\label{fig:case1}
\end{figure}

We conducted another type of simulation by supposing we have input of 20 tokens as well as an identical amount of output.
We simulated two simultaneous translation strategies, wait-k and fixed-size segmentation.
In the fixed-size segmentation, we assume for simplicity the length of the input and output chunks to be $k$ until the end-of-sentence token is predicted.
We call this simple strategy chunk-$k$ in this simulation.
Hyperparameter $k$ for wait-$k$ and chunk-$k$ varies from 1 to 20.

\autoref{fig:case1} shows the results.
Except for AL evaluating chunk-$k$, since all the metrics have identical values, they are plotted using the same color.
There is a gap between wait-$k$ and chunk-$k$ when using AL, although ATD and DAL result in the same values for them.
One serious problem raised here is the large jump in AL at $k=20$.
For chunk-$19$, AL uses $r=1$, $\tau_{g_{\textrm{chunk-}19}}(|\bm{x}|) = 20$, $g_{\textrm{chunk-}19}(\tau)=19$ ($1 \leq \tau \leq 19$), and $g_{\textrm{chunk-}19}(\tau)=20$ ($\tau = 20$).
The AL value becomes $\frac{1}{20} \left\{ \left( \sum_{\tau=1}^{19} \tau \right) + 1 \right\} = \frac{191}{20} = 9.55$.
However, in the case of chunk-$20$, $g_{\textrm{chunk-}20}(\tau) = 20$ for all $\tau$ and  $\tau_{g_{\textrm{chunk-}20}}(|\bm{x}|) = 1$, according to \autoref{eqn:tau_g}.
As a result, the AL value becomes 20.
This phenomenon shows the problem of AL in which a long partial translation output reduces the latency.

\subsection{Case 4: (10+10)-($L_1$+10)}\label{subsec:case4}

\begin{figure}[t]
\centering
\centerline{\includegraphics[width=8.5cm]{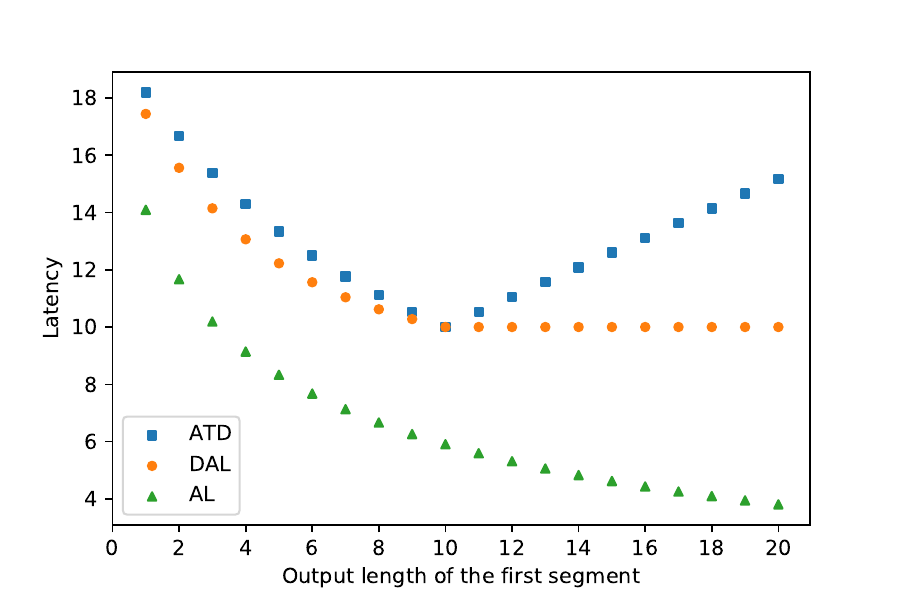}}
\caption{Case 4 (10+10 input and $L_1$+10 output tokens with varying $L_1$)}
\label{fig:case4}
\end{figure}

In another case, we assume the input is divided into two segments, each of which has 10 tokens and each corresponding output consists of $L_1$ and 10 tokens with varying $L_1$.
\autoref{fig:case4} shows the results.

If $L_1 < 10$, some target tokens corresponding to the first input chunk come after the start of reading the second chunk.
\autoref{tab:equation2b} illustrates such a situation, in which $y_2$ is in the second chunk output although the corresponding $x_2$ is in the first chunk input.
If $y_2$ is in the first chunk output, the time difference between $x_2$ and $y_2$ becomes smaller. It also reduces the time difference of the latter token pairs, such as  $x_3$ and $y_3$.
Therefore, the shorter $L_1$ becomes, the larger the delay grows in this range.

If $L_1 > 10$, the translation of the second chunk is delayed due to the long translation outputs for the first chunk (\autoref{tab:equation2c}).
Therefore, the longer $L_1$ becomes, the larger the delay grows in this range. ATD reflects this phenomenon, but AL decreases monotonically with $L_1$; DAL does not increase.

\section{ SimulMT Analyses }
We conducted analyses on actual SimulMT results to investigate ATD's effectiveness.
Since most existing latency metrics were originally proposed for text-to-text SimulMT, we compared text-to-text models in this experiment.

\subsection{Data}
We used the data from the shared task of English-to-German simultaneous translation in the IWSLT 2022 evaluation campaign \cite{anastasopoulos-etal-2022-findings}.
We used the WMT 2014 training set (4.5 M sentence pairs) for pre-training and the IWSLT 2017 training set (206 K sentence pairs) for fine-tuning.
The development set consisted of dev2010, tst2010, tst2011, and tst2012 (5,589 sentence pairs in total), and the evaluation set was tst2015 (1,080 sentence pairs).

Following the experimental settings in the literature \citep{kano-etal-2022-simultaneous}, we compared several SimulMT methods: wait-$k$ \citep{ma-etal-2019-stacl}, Meaningful Unit \citep[MU;][]{zhang-etal-2020-learning-adaptive}, Incremental Constitute Label Prediction \citep[ICLP;][]{kano-etal-2021-simultaneous}, and Prefix Alignment \citep[PA;][]{kano-etal-2022-simultaneous}.

\subsection{Results}

\begin{figure}[ht]
\begin{minipage}[b]{1.0\hsize}
\centering
\centerline{\includegraphics[width=7.0cm]{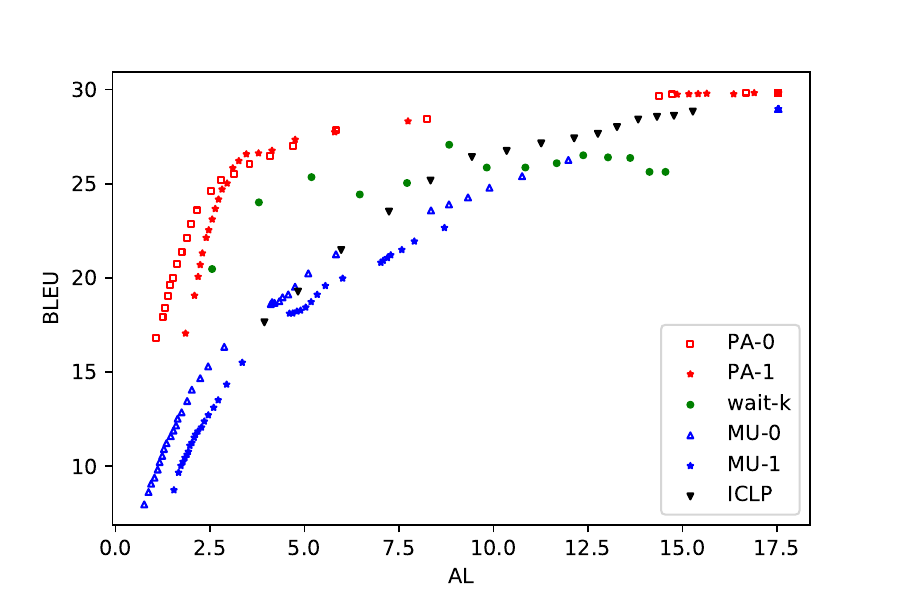}}
\vspace{-2mm}
\subcaption{Latency measurement by AL}\label{fig:result_de_al}
\end{minipage}

\begin{minipage}[b]{1.0\hsize}
\centering
\centerline{\includegraphics[width=7.0cm]{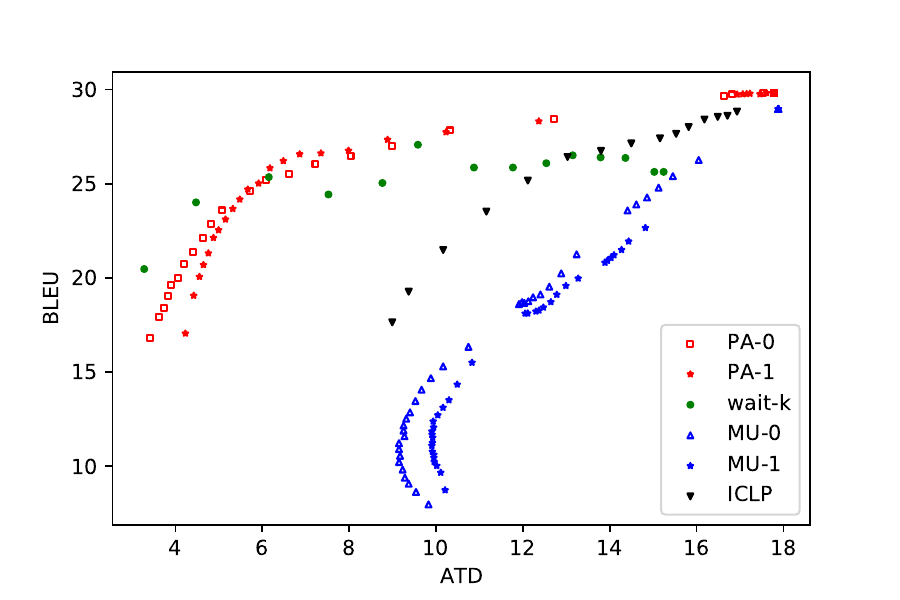}}
\vspace{-2mm}
\subcaption{Latency measurement by ATD}\label{fig:result_de_atd}
\end{minipage}
\caption{Latency and BLEU}\label{fig:en-de}

\centerline{\includegraphics[width=7.0cm]{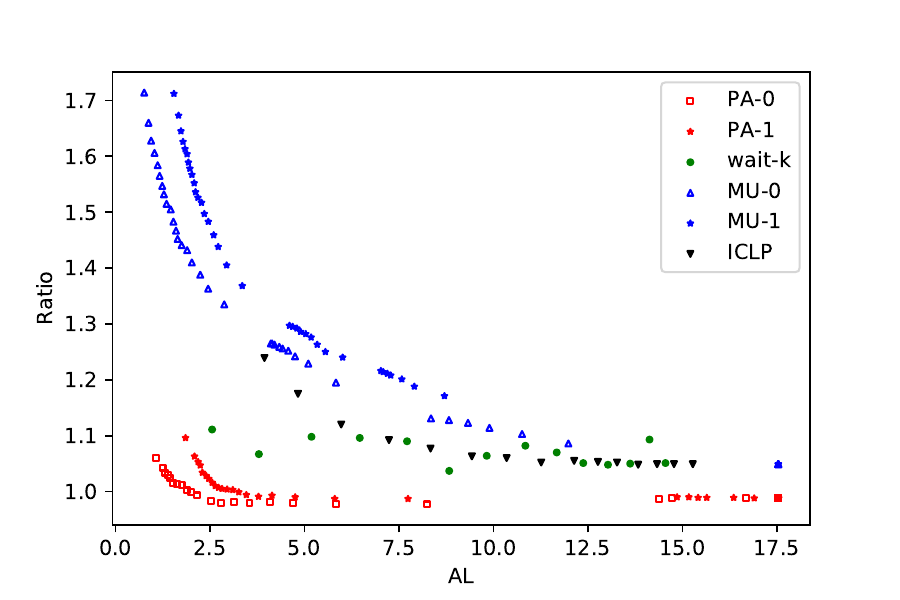}}
\vspace{-4mm}
\caption{Length ratio and AL}\label{fig:ratio_de_al}
\end{figure}

\autoref{fig:en-de} shows the results.
Compared to the AL shown in \autoref{fig:result_de_al}, the ATD in \autoref{fig:result_de_atd} demonstrated clear differences in delay among models.
MU and ICLP were affected by the change in the latency metric.
We scrutinized their results and found this degradation was caused by over-translation, suggested by observations of the length ratio results shown in \autoref{fig:ratio_de_al}.
MU and ICLP generated long translations exceeding a length ratio of 1.0 when they worked with small latency.
One interesting finding here is the correlation between BLEU and ATD by MU;
a larger latency did not always increase BLEU. Over-translation increased ATD, but it simultaneously decreased BLEU.
In contrast, the wait-$k$ strategy just generates one output token at a time and does not suffer from this issue.
PA also worked well with the latency measurement by ATD because it fine-tunes the translation model to prevent over-translation.

\section{Using EVS as reference}
Since the above analyses were not enough to verify ATD's effectiveness, we conducted the following experiments to compare different latency metrics.
To verify which was superior, we compared the correlation between each latency metric and the mean EVS, as calculated from the outputs of real speech-to-speech SimulMT.

\subsection{Methodology to calculate EVS}

EVS can be measured in several ways \cite{elisa2019}.
We used an automatic word alignment tool to assist humans to choose the correct alignment due to the high cost of word alignment from scratch by humans.
EVS is calculated by the following process:

\paragraph{1.} Run speech-to-speech SimulMT models for the test dataset and output target speech.
\paragraph{2.} Use awesome-align \cite{dou2021word} to obtain automatic word alignment in each sentence transcription of the source and target speech. Transcription is obtained by Automatic Speech Recognition (ASR) using Whisper \cite{radford2022robust}.
\paragraph{3.} A human annotator chooses the correct alignment from the awesome-align output.
\paragraph{4.} Obtain the timestamps for the source and target words from the speech and its transcription by WhisperX \cite{bain2022whisperx}.
\paragraph{5.} Take the time difference between the start time of the source and target words with a correct alignment as EVS, and calculate the mean value of EVSs in a sentence as the mean EVS. 

For the analysis, we also calculated the mean value of all the EVSs, including the wrong word alignments and call this the mean automatic EVS.

\subsection{Settings}
We used the following models and evaluation dataset in our experiment.
\subsubsection{Models}

We used a cascaded speech-to-speech SimulMT model by connecting the speech-to-text SimulMT and incremental Text-To-Speech (TTS) models following \citet{fukuda-etal-2023-naist}.
We used their TTS model \citep{fukuda-etal-2023-naist} for the following three types of speech-to-text SimulMT models and their variations:

\paragraph{Test-time wait-k \citep{ma-etal-2019-stacl}:} This method first reads \texttt{k} input speech segments and alternately repeats outputting one word and reading one input speech segment.
Hyperparameter \texttt{k} of test-time wait-k was set to 3 and 5. 
The source speech is input to the translation model in 250-ms segments. 

\paragraph{Fixed-size segmentation:} This method immediately starts a translation every time a new input speech segment comes.
We compared segment sizes of 750 and 2500 ms.

\paragraph{System by \citet{fukuda-etal-2023-naist}:} 
This scheme uses Prefix Alignment \citep{kano-etal-2022-simultaneous} and Local Agreement \citep{liu2020lowlatency} to mitigate over-translation with fixed-size segmentation.
We compared segment sizes of 250 and 1250 ms.

The settings of the training NMT models followed published research \citep{fukuda-etal-2023-naist}.
For fixed-size segmentation and test-time wait-k, we shared a single NMT model, which was not fine-tuned with prefix alignment pairs.

\subsubsection{Evaluation data}
We used the first 30 sentences of the English-Japanese tst-COMMON in MuST-C.
We used three types of SimulMT methods, each of which has two variations as described above.
180 sentences are evaluated.

In the streaming SimulMT, since there are no obvious sentence boundaries, the input sequence sometimes becomes longer than a sentence.
Such a long translation can delay the translation of subsequent inputs.
Therefore, we also evaluated the latency of the concatenation of two sentences.

We used Spearman's $\rho$ to measure the correlation between EVS and each latency metric.

\subsection{Compared latency metrics}
We compared ATD with the baselines of the speech-to-speech and text-to-text latency metrics.
\subsubsection{Speech-to-speech}
Since the existing metrics described in Section 3 cannot be applied to speech-to-speech SimulMT, we used two baseline latency metrics for this experiment: Start Offset and End Offset, both of which were officially used in the Simultaneous Translation Task of IWSLT 2023 Evaluation Campaign \citep{agrawal-etal-2023-findings}.
Start Offset is literally the time difference between the starts of the source speech input and the target speech output.
End Offset uses the ends instead of the starts.
For ATD, we set $\tau$ to 300 ms in the experiment.

We evaluated the latency metrics using the speech-to-speech SimulMT outputs and
conducted an experiment on both computation-aware (CA) and non-computation-aware (NCA) conditions.
In the former, the output speech reflected the actual computation time as silence.
Therefore, the target token timestamps in the CA condition should be different from those in the NCA condition.
We used one NVIDIA RTX TITAN throughout the experiments.

\subsubsection{Text-to-text}
We compared AL \cite{ma-etal-2019-stacl}, AL with reference \cite{ma-etal-2020-simuleval}, LAAL \cite{papi-etal-2022-generation}, DAL \cite{arivazhagan-etal-2019-monotonic}, and ATD on the text-to-text condition.
By using the word timestamps obtained by WhisperX, we corresponded the chunk boundaries to the text boundaries.

\subsection{Result: speech-to-speech}
\begin{table*}[t]
\centering
\begin{tabular}{l|cc|cc}
\hline
\multicolumn{1}{l}{} & \multicolumn{2}{|c|}{1 sentence} & \multicolumn{2}{c}{2 sentences} \\
\hline
Metris & NCA & CA & NCA & CA \\
\hline
Start Offset & 0.632 (0.000) & 0.374 (0.000) & 0.368 (0.000) & 0.103 (0.225) \\
End Offset & 0.611 (0.000) & 0.755 (0.000) & 0.541 (0.000) & 0.755 (0.000) \\
ATD & \textbf{0.836} (0.000) & \textbf{0.859} (0.000) & \textbf{0.779} (0.000) & \textbf{0.832} (0.000) \\
\hline
Mean auto. EVS & 0.869 (0.000) & 0.897 (0.000) & 0.839 (0.000) & 0.887 (0.000) \\
\#samples & 149 & 148 & 143 & 142 \\
\hline
\end{tabular}
\caption{Speech-to-speech: Spearman's correlation between mean EVS and latency metrics: Value in parentheses is p-value.}
\label{tb:s2s_human_res}
\end{table*}

\begin{table}[t]
\centering
\begin{tabular}{l|c|c}
\hline
\multicolumn{1}{l}{Metrics} & \multicolumn{1}{|c|}{1 sentence} & \multicolumn{1}{c}{2 sentences} \\
\hline
AL & 0.568 (0.000) & 0.246 (0.003) \\
AL-ref & 0.602 (0.000) & 0.257 (0.002) \\
LAAL & 0.624 (0.000) & 0.270 (0.001) \\
DAL & 0.646 (0.000) & 0.381 (0.000) \\
ATD & \textbf{0.651} (0.000) & \textbf{0.461} (0.000) \\
\hline
\#samples & 149 & 143 \\
\hline
\end{tabular}
\caption{Text-to-text: Spearman's correlation between mean EVS and latency metrics}
\label{tb:t2t_human_res}
\end{table}

\begin{table}[t]
\centering
\begin{tabular}{lc}
\hline
\# Sentence pairs & 223,108 \\
\# English words & 4,593,204 \\
\# Japanese words & 4,794,912 \\
\# Japanese characters & 8,838,777 \\
\hline
\end{tabular}
\caption{Statistics of IWSLT2017 English-Japanese training dataset}
\label{tb:char_word_num}
\end{table}

\begin{table*}[t]
\centering
\begin{tabular}{l|cc|cc}
\hline
\multicolumn{1}{l}{} & \multicolumn{2}{|c|}{1 sentence} & \multicolumn{2}{c}{2 sentences} \\
\hline
Metric & 1 char & 2 char & 1 char & 2 char \\
\hline
AL & 0.567 (0.000) & 0.566 (0.000) & 0.149 (0.075) & 0.147 (0.080) \\
AL-ref & 0.575 (0.000) & 0.613 (0.000) & 0.193 (0.021) & 0.264 (0.001) \\
LAAL & 0.613 (0.000) & 0.614 (0.000) & 0.198 (0.018) & 0.264 (0.001) \\
DAL & \textbf{0.634} (0.000) & 0.630 (0.000) & 0.302 (0.000) & 0.313 (0.000) \\
ATD & 0.601 (0.000) & \textbf{0.641} (0.000) & \textbf{0.498} (0.000) & \textbf{0.506} (0.000)\\
\hline
\#samples & 149 & 149 & 143 & 143\\
\hline
\end{tabular}
\caption{Text-to-text (character-level output): Spearman's correlation between mean EVS and latency metrics}
\label{tb:t2t_human_res_char}
\end{table*}

Table~\ref{tb:s2s_human_res} shows the correlation between the mean EVS and latency metrics for the speech-to-speech SimulMT results.
\#samples represent the number of evaluated sentences after removing the sentences with no correct word alignments.
Regardless of the input sequence length and the computation-awareness, ATD has the highest correlation in all the latency metrics that do not use word alignment.
Mean automatic EVS is the best because the mean EVS is calculated based on word alignments, which are part of the alignments used to calculate the mean automatic EVS.

For the two sentences with both CA and NCA, the correlation of Start Offset largely decreased from the result for one sentence.
When the input increased, the impact of the delay at the first start of the translation fell for the entire delay.
Therefore, Start Offset has lower correlation, and in the CA condition, it has no significant correlation.
ATD maintains a relatively high correlation, as does the mean automatic EVS.
When the input sequence is lengthened, the accumulated delay caused by the previous long chunk translation output increases.
Since ATD adequately considers the output length, it still has high correlation.

\subsection{Result: text-to-text}

Table~\ref{tb:t2t_human_res} shows the correlation between the mean EVS and text-to-text latency metrics. 
ATD has the highest correlation among all the text-to-text latency metrics.
AL is the most common latency metric, but its correlation was worse than ATD and its variants (AL-ref, LAAL, and DAL).
DAL has the best correlation among the AL variants.

For two sentences, the correlations of all the latency metrics greatly decreased, but the differences between ATD and the other latency metrics are larger than for one sentence.
This is because ATD includes enough delay caused by the end time of the previous sentence translation, as explained in the speech-to-speech result.

\subsection{Result: text-to-text (character-level)}
Text-to-text latency metrics basically assume each text token is a word. 
However, languages like Japanese and Chinese have no white spaces between words. 
Therefore, SimulEval \cite{ma-etal-2020-simuleval}, which is commonly used to evaluate simultaneous translation systems, regards characters as tokens for Japanese for simplicity without tokenizing them into words.
We additionally conducted an evaluation with character-based text tokens.

\autoref{tb:char_word_num} shows the statistics of the IWSLT 2017 English-Japanese training dataset.
The number of English and Japanese words is close, although the number of Japanese characters is twice as large.
Therefore, to calculate the latency metrics, we compared one character as one token and two characters as one token \footnote{For two characters as one token, we regard the last one character in the output chunk as one token if the number of characters in the chunk is odd.}.

\autoref{tb:t2t_human_res_char} shows the result.
Basically, the word-level latency calculation in \autoref{tb:t2t_human_res} has higher correlation than the character-level, and two characters as one token is better than one character as one toke, especially for two sentences.
This is because the amount of information conveyed by one Japanese word is closer to that of one English word than one or two Japanese characters.
For the same reason, two Japanese characters as one token outperformed one Japanese character as one token, according to \autoref{tb:char_word_num}.

ATD is the best except for a condition of one character as one token for one sentence.
ATD is the latency metric most affected by the output length.
An output of one Japanese character has less correspondence with the input of one English word than other output-levels, and the output length by character is much longer than the input length by word.
As a result, ATD added excessive delay caused by this long output, and its correlation decreased.

According to the character-level result, we must carefully address tokenization, not only for ATD, but also for other latency metrics.

\section{Conclusion}

We proposed a novel latency metric ATD for SimulMT, which addresses the problem of latency evaluations for long chunk outputs by taking the output length into account.
ATD gives a large latency value to a long output based on the assumption that the output also causes a delay, unlike in AL.
We identified ATD's effectiveness by analyzing the simulations and it had the highest correlation with EVS among the token-based latency metrics.
Despite being much simpler than mean automatic EVS and without requiring a long process like ASR, word alignment, or getting timestamps, ATD correlates mean EVS very well.
Therefore, it can be easily used to evaluate the latency of simultaneous translation.
In future work, we will investigate the correlation between latency metrics and the delay experienced by listeners. 

\section{Acknowledgement}
This paper is an extended version of our conference paper \cite{kano2023average} with additional experiments and analyses.
Part of this work was supported by JSPS KAKENHI Grant Numbers JP21H05054 and JP21H03500.

\bibliography{tacl2021,anthology,custom}
\bibliographystyle{acl_natbib}

\iftaclpubformat

\onecolumn

\appendix





  

\end{CJK}
\end{document}

I suggest expanding this idea a bit because this description is rather vague.
Isn't this what you mean?
For what it's worth, your example English sentence here is slightly awkwardl: gI buy a pen.h The simple present tense would never be used for this situation or context. Either gI'm buying a penh or gI will buy a pen.h
Unclear.
Very unclear how ginput wordsh become available? Are you just referring to the literal number of words in a chunk? 
Isn't this what you mean?
Unclear what exactly is being compared here.
[i8]Commonly used what?
[i9]Awkward use of gash. Check my revisions carefully.
[i10]In Fig. 2, remove the extra space before colons when necessary. Also remove the capital letter on Output.
[i11]Awkward use of gsimplifyh here. Do you mean that you made the durations the same? We made the duration lengths identical?
[i12]I suggest tightening this to gFigs. 4(a) and (b)h.
[i13]Right?
[i14]Unclear what this verb means in this context.
[i15]Unclear. Does this verb (spend) mean use or consume here?
[i16]Best at what? 
[i17]Unclear.
[i18]Best what? Is this what you mean? Be specific.
[i19]Best what?